\documentclass[letterpaper]{article} 
\usepackage{aaai25}  
\usepackage{times}  
\usepackage{helvet}  
\usepackage{courier}  
\usepackage[hyphens]{url}  
\usepackage{graphicx} 
\usepackage{natbib}  
\usepackage{caption} 
\usepackage{algorithm}
\usepackage{algorithmic}
\usepackage{amsmath}
\usepackage{amsfonts}
\usepackage{multirow}
\usepackage{newfloat}
\usepackage{listings}
\DeclareCaptionStyle{ruled}{labelfont=normalfont,labelsep=colon,strut=off} 
\floatstyle{ruled}
\newfloat{listing}{tb}{lst}{}
\floatname{listing}{Listing}
\title{EDENet: Echo Direction Encoding Network for Place Recognition Based on Ground Penetrating Radar}
\author{
    Pengyu Zhang\textsuperscript{1,2}, 
    Xieyuanli Chen\textsuperscript{1}, 
    Yuwei Chen\textsuperscript{1}, Beizhen Bi\textsuperscript{1}, Zhuo Xu\textsuperscript{1}, Tian Jin\textsuperscript{1}, \\Xiaotao Huang\textsuperscript{1}, Liang Shen\textsuperscript{\rm 1 }\thanks{Corresponding authors}\\
}
\affiliations{
    \textsuperscript{\rm 1}National University of Defense Technology, Changsha, 410073, China\\
    \textsuperscript{\rm 2}National University of Singapore, 117576, Singapore\\

    $\{$zhangpengyu19, xieyuanli.chen, yuweichen15, beizhenbi20, xuzhuo19, tianjin, xthuang, liangshen16$\}$@nudt.edu.cn
}

\usepackage{bibentry}

\begin{document}

\maketitle

\begin{abstract}
Ground penetrating radar (GPR) based localization has gained significant recognition in robotics due to its ability to detect stable subsurface features, offering advantages in environments where traditional sensors like cameras and LiDAR may struggle. However, existing methods are primarily focused on small-scale place recognition (PR), leaving the challenges of PR in large-scale maps unaddressed. These challenges include the inherent sparsity of underground features and the variability in underground dielectric constants, which complicate robust localization.
In this work, we investigate the geometric relationship between GPR echo sequences and underground scenes, leveraging the robustness of directional features to inform our network design. We introduce learnable Gabor filters for the precise extraction of directional responses, coupled with a direction-aware attention mechanism for effective geometric encoding. To further enhance performance, we incorporate a shift-invariant unit and a multi-scale aggregation strategy to better accommodate variations in dielectric constants.
Experiments conducted on public datasets demonstrate that our proposed EDENet not only surpasses existing solutions in terms of PR performance but also offers advantages in model size and computational efficiency. 

\end{abstract}

%

\section{Introduction}

Ground Penetrating Radar (GPR) for vertical detection of robust and invariant underground features has garnered significant attention in recent years in vehicle/robot localization \cite{RF20,RF10,RF4}. 
The straightforward idea for this task is to use the GPR scans to perform scan matching in the prior map like LiDAR \cite{RF6,RF7}.
Due to the limited observation range of a single GPR, some work attempts to combine IMU to improve the performance \cite{RF9}.
Multiple weather scenarios can cause underground dielectric constants to change, which in turn changes echo characteristics.
Some recent work has attempted to overcome this limitation through deep networks \cite{RF10, RF25}, but it is not robust to interference noise.
In order to meet the needs of real-time application, DEC was developed to compress the original echo sequence, and achieve better results than deep networks in local area place recognition (PR), but the DEC features lack uniqueness and have a high mismatch rate on large-scale maps \cite{RF1}.

Using GPR for PR presents several critical limitations.
First, unlike the abundant features available for cameras and LiDAR, underground features are limited. Directly adapting existing methods to GPR often results in redundant parameters and convergence difficulties.
Second, radar echoes are highly prone to interference, yet most deep learning models rely on CNNs with fixed weights. This uniform convolutional encoding amplifies the network's sensitivity to noise and interference.
Third, variability in Dielectric Properties: Multi-temporal GPR data is influenced by fluctuations in underground dielectric constants, causing significant variability in extracted features. Current methods fail to effectively address the challenge of robust feature extraction under these changing conditions.

This paper proposes EDENet, a novel approach for generating compact sequential descriptors through synchronized feature encoding and filtering representation design. 
Specifically, we propose the incorporation of learnable Gabor filters to integrate geometric priors from GPR echoes, thereby constraining the feature encoding process of the network. 
Furthermore, the implementation of a direction-aware attention mechanism enables adaptive geometric encoding and feature filtration. 
Finally, by employing a multi-scale design and cascading existing feature aggregation layers, our approach achieves state-of-the-art performance.

In sum, the contribution of this article is summarized as follows:
\begin{itemize}
    \item In the scenario of underground dielectric constant changes, the robust directional features in underground echoes are analyzed for the first time to guide the design of the network.
    \item A method based on learnable Gabor filter layers and direction-aware attention is proposed to adaptively encode robust directional features and filter out useless information. 
    At the same time, EDEBlock is constructed by integrating offset-invariant units.
    \item Based on multi-scale EDEblock, a new network EDENet is proposed for GPR-based place recognition. 
    Integration with the existing simple feature aggregation layer can achieve impressive performance.
    Extensive quantitative and qualitative experiments not only verify the performance, but also show that the EDENet has obvious advantages in model size and computing speed. Our codes are avaliable\textsuperscript{\rm 1 }\thanks{{https://github.com/Pyxel0524/EDE-LGPR.}
    }
\end{itemize}

\section{Related Works}
\label{gen_inst}
\subsection{Underground Scene Representing}
\textbf{Handicraft Descriptors} often combine near-field electromagnetic reflection characteristics to capture ripple features in GPR images. 
For example, the Dense Energy Curve (DEC) has been applied to GPR-based localization, but it lacks unique discriminative ability, leading to high false alarm rates in PR tasks \cite{RF1}. 
Some methods emphasize hyperbolic features of significant underground targets based on geometric relationships \cite{RF15,RF17}, but these are limited by GPR signal attenuation\cite{RF47,RF48}. Log-Gabor filtering in the frequency domain has been explored for target detection \cite{RF13, RF19}, yet handcrafted features generally require manual adjustment of filtering parameters, which is time-consuming and dataset-dependent.

\textbf{Learning-based representation} leverage deep networks for specific tasks. 
In underground target detection, CRNet \cite{RF30} uses a UNet-based structure to filter clutter and reconstruct features, while M2FNet employs multi-view learning for feature identification \cite{RF24}. 
CMU-Net, based on ResNet, overcomes clutter by estimating transformations between GPR slices \cite{RF9}, and LGPRNet uses Inception modules to extract multi-scale features for robot position regression \cite{RF10}.
In comparison, MWSNet developed based on Siamese-UNet shows better performance in overcoming location recognition tasks under different weather conditions \cite{RF25}.
However, the above CNN-based approaches excel in handling large datasets but are sensitive to redundant details from dielectric constant variations, leading to overfitting. 
Our approach aims to automatically filter unnecessary details and model directional responses at multiple scales for more robust global representation.

\subsection{Place Recognition (PR)}

Previous research in place recognition has predominantly relied on surface-based sensors, such as cameras and LiDAR. 
The strategy of compressing raw sensor observations into compact descriptors, as exemplified by methods like NetVLAD \cite{RF33,RF31} for vision and OverlapNet \cite{RF44} for LiDAR, has been widely adopted to accelerate localization. 
To better account for the continuity of robotic motion, recent advancements, including SeqVLAD and SeqOT \cite{RF21,RF22}, have focused on incorporating sequential information \cite{RF32}, thereby improving upon their respective single-observation counterparts.

However, while these methods have demonstrated success in feature-rich surface environments, they face significant challenges in the feature-sparse underground settings. 
This highlights the need for a targeted analysis to identify effective features for underground environments.

\section{Motivation}
In this section, we analyze the key characteristics of GPR echoes, which motivate the design of our proposed method.
GPR detects changes in the dielectric constant of subsurface materials by emitting electromagnetic waves.
Due to the continuity of the underground medium, significant subsurface features often display directional geometric properties in the resulting echo sequences\cite{RF47}. 
Assuming a homogeneous dielectric constant $\epsilon_r$ and considering GPR movement along a trajectory distance $x$, beamwidth $\theta$, and electromagnetic wave speed $c$, the GPR echoes can be approximated by a hyperbolic form, as shown in Eq.\ref{eq:echo}.

\begin{figure}[htbp]
    \centering
    \includegraphics[width=1\linewidth]{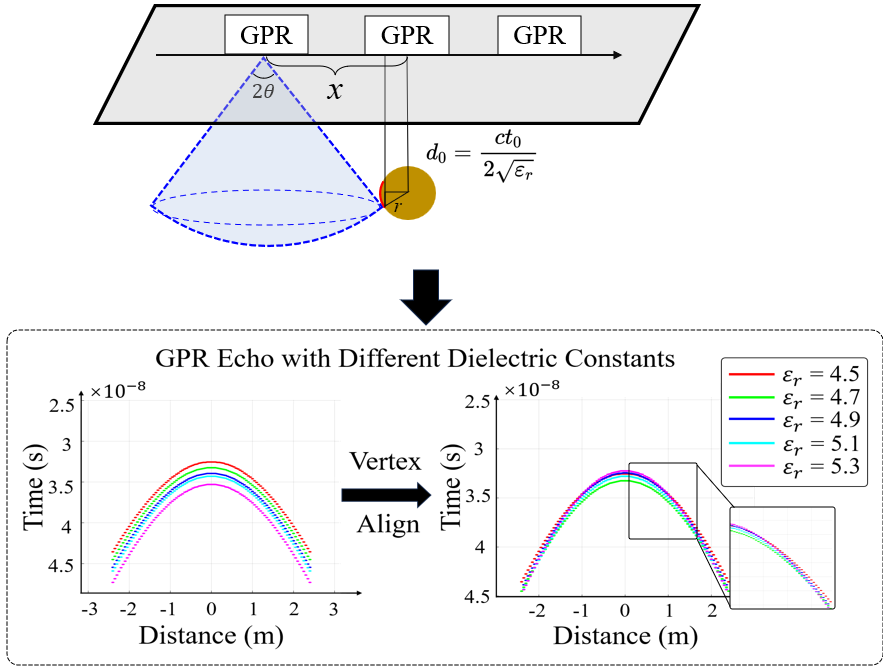}
    \caption{GPR echoes with different dielectric constants.
    }
\label{detect}
\end{figure}

\begin{figure}[htbp]
    \centering
    \includegraphics[width=0.5\linewidth]{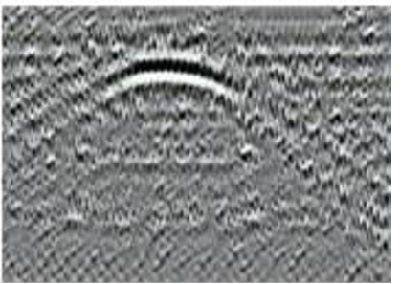}
    \caption{Curve features obscured by interference.
    }
\label{noise}
\end{figure}

\begin{equation}
\label{eq:echo}
\left\{\begin{array}{l}
{\left[x-\left(\frac{c}{2 \sqrt{\varepsilon_r}}\right) \sin \theta\right]^2+\left[\left(\frac{c}{2 \sqrt{\varepsilon_r}}\right) \cos \theta-\left(r+d_0\right)\right]^2} \\ {=r^2 \quad \quad \quad \quad  \quad \quad\quad \quad \quad \quad \quad \frac{(d+r)}{cos\theta } <\left | x \right | <\frac{(d+2r)}{cos\theta } } \\

\frac{\left(\frac{c t}{2 \sqrt{\varepsilon_r}}+1\right)^2}{\left(\frac{d_0}{r}+1\right)^2}-\frac{x^2}{\left(d_0+r\right)^2}=1 \quad \quad \quad \quad \quad \quad \quad \quad others, \\

\end{array}\right.
\end{equation}

\begin{figure*}[htbp]
    \centering
    \includegraphics[width=1\linewidth]{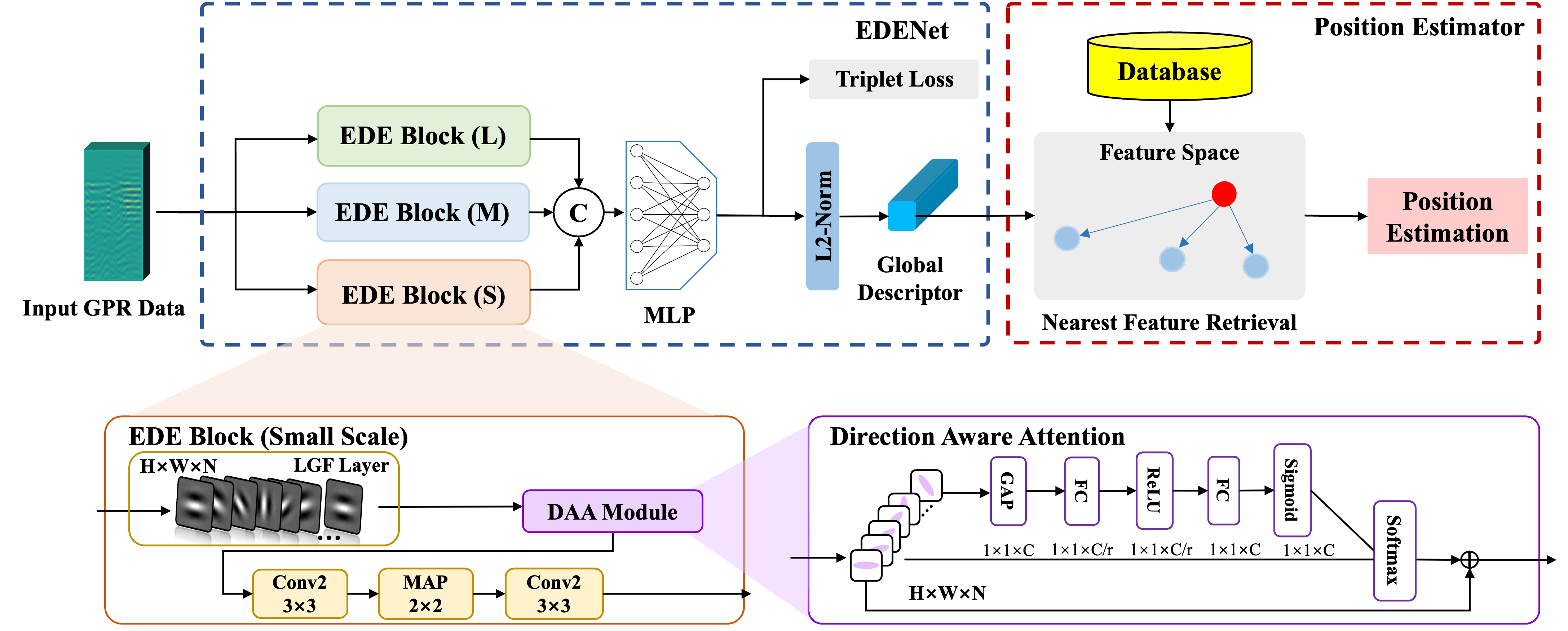}
    \caption{The proposed framework consists of two modules. In EDENet, the sequential GPR data is compressed to generate a compact descriptor. In the position estimator module, we use a vector retrieval scheme to find the best matching.
    }
\label{framework}
\end{figure*}

For radar echoes collected by the same GPR, $\epsilon_r$ is the main factor affecting the radar image. 
Variations in $\epsilon_r$ alter the propagation speed of the waves, leading to feature distortions and shifts along the depth axis. 
However, as shown in Fig. \ref{detect}, when aligning echoes across different dielectric constants, the geometric features remain relatively consistent, while shifts vary. Therefore, capturing the directional information in the echoes is crucial for encoding the GPR sequence's features.

On the other hand, as illustrated in Fig. \ref{noise}, electromagnetic interference and noise in extreme conditions introduce random perturbations that can obscure these directional features. Thus, a feature selection model that integrates feature encoding with noise filtering is essential for generating a compact and robust descriptor.

\section{Methodology}

\subsection{Problem Definition}
In this study, we address the challenge of PR based on underground sequence echoes. 
To achieve this goal, we define a sequence descriptor retrieval problem for prior map data $\mathcal{M} = \left\{(\bold {m}_i, \bold {T}_i),i=1,2,3,...,n\right\}$.
The database formed by $n$ GPR echo slices $m_i$ and their associate UTM pose $\bold {T}_i = (\bold {utm}_{x}^{i}, \bold {utm}_{y}^{i})$, where each slices $m\in \mathbb{R}^{D\times C}$, with $D$ representing the dimension of depth bins and $C$ representing the dimension of radar array channels, corresponding to the range and azimuth information of the radar echo, respectively.
Specifically, $\bold {m}_{i,D,C}$ represents the intensity of the radar return in the $i$th frame at a specific depth and channel.
Additionally, each frame $m$ is associated with a pose that contains the pose of the RTK GPS.
Next, we have a sequence of query frames $\left\{\bold {q}_i,\bold {q}_{i+1},...,\bold {q}_{i+n}\right\}$, measured when the vehicle revisits the map. 
Our goal is to find the optimal correspondence of $\left\{\bold {m}_i,\bold {m}_{i+1},...,\bold {m}_{i+n}\right\}$in $\mathcal M$ which provides the position at the time $\bold m$ was measured.

\subsection{Architecture Overview}

EDENet encodes directional information from radar echo sequences using multi-scale EDEblocks to construct a geometric representation.
Each EDEblock consists of a series of learnable Gabor filters for capturing directional responses and a Directional Attention Module for adaptive reweighting, enhanced with a shift-invariant unit for dimension reduction. 
Multi-scale features finally are aggregated using an MLP at the output layer to generate compact global descriptor.
This design allows EDENet to geometrically encode features while filtering out noise and irrelevant information effectively.

\subsection{Echo Direction Encoding (EDE) Block}
\subsubsection{Learnable Gabor Filters (LGF) Layer}
\label{LGF layer}
To extract robust underground features, several convolution operators have been mentioned in the past in underground target detection and image processing tasks, including Gaussian Filter \cite{RF34},  Gabor filter \cite{RF43}, and Laden transform \cite{RF27}.
Inspired by the application of Gabor features in geology and GPR-based tasks \cite{RF41,RF42}, this paper proposes a learnable Gabor  ($\mathcal LG$) filter and constructs a $\mathcal LG$-based layer (LGF).
The LGF is defined as follows:
\begin{equation}
\label{eq: learnable gabor}
\begin{split}
\mathcal LG&\left ( i,j;\lambda, \gamma,\phi ,\theta ,\sigma \right ) = \\
&-exp\left \{ -\frac{i'^{2}\gamma ^{2}+ j'^{2}}{2\sigma ^{2} }  \right \}cos\left (2\pi \frac{i'}{\lambda }+\phi    \right ),  
\end{split}
\end{equation}
the pixel indices are denoted by $i$ and $j$, and $i'=icos\theta +jsin\theta$, $j'=-isin\theta +jcos\theta $.
$\lambda$ represents the wavelength of the sinusoidal wave component, $\gamma$ defines the ellipticity of the Gaussian support of the Gabor filter, $\phi$ is the phase shift, and $\theta$ and $\sigma$ determines the direction and the standard deviation of the basic wave.
All these parameters control the shape of filter.

We further remove the zero DC component by subtracting the mean, which allows the Gabor filter to approximate the compact support property \cite{RF35}:
\begin{equation}
\label{eq:gabor in EDE}
\begin{split}
\mathcal F&\left ( i,j,K \right )=\mathcal LG\left (i-\frac{K}{2},j-\frac{K}{2} \right ) \\
&-\frac{1}{K^2} \sum_{x= 0}^{K} \sum_{y=0}^{K} \mathcal LG\left (i-\frac{K}{2},j-\frac{K}{2} \right ) ,
\end{split}
\end{equation}

As illustrated in Eq. \ref{eq: learnable gabor}, when $\lambda$ tends to $\infty$ and $\phi$ approaches 0, the learnable Gabor filter transforms into a Gaussian filter kernel. 
When $\lambda$ tends to $\infty$ and $\gamma$ and $\phi$ approach 0, the learnable kernel becomes a edge fitting filter, which is used for edge direction fitting in radar images \cite{RF27}. 
Thus, the learnable Gabor filter can represent various types of filters.
It is capable of extracting valuable directional responses and can also act as a filter to remove irrelevant information and clutter from the image.

\begin{figure}[htbp]
    \centering
    \includegraphics[width=1\linewidth]{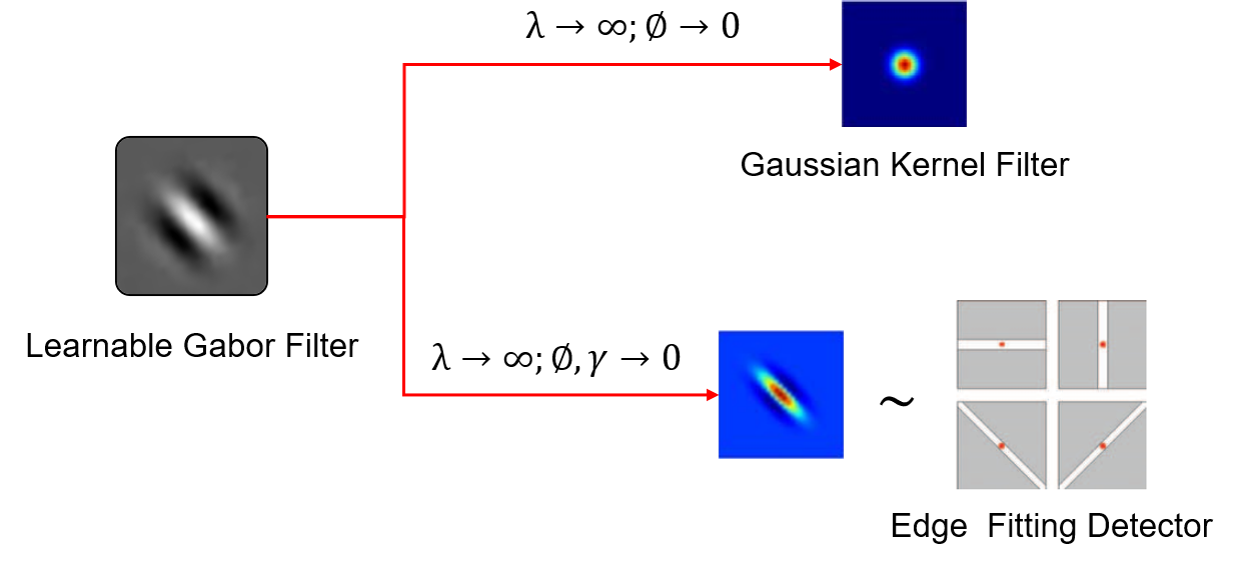}
    \caption{The learnable Gabor filter.
    }
\label{gaborfilter}
\end{figure}

\textbf{Analyze}: The LGF module is designed to extract multi-direction responses.
Experiments have found that overly freedom parameter learning will overfitting.
Such as setting $\theta$ as a learnable parameter will cause the network to focus only on specific main directions.
Therefore, in this paper, the LGF module is defined as composed of a set of LG $\left \{ \mathcal F(i,j,K;\theta_k)|k=1,2,3,...,n \right \}$, the $k$-th $\theta$ is defined as $\theta_k = k\cdot\pi/n$, each LGF module shares the same $\sigma$ and receptive field $K$.
This ensures the diversity and flexibility of the LG modality.

\subsubsection{Direction Aware Attention (DAA)}
\label{DAA module}
Our goal is to ensure that the network can improve sensitivity to specific directional filtering and suppress less useful filtering kernels. 
To achieve this goal, we suggest modeling the interdependence of kernels $\mathcal F$, the weights of filters in different directions are recalibrated through two steps, first by modeling the interdependence of direction kernels and then by incentivizing. 
In the phase of modeling, the output of LGF module $\bold V=\sum_{k=1}^{n}\mathcal F_{k}\ast \bold{X}$ as a local spatial representation of GPR sequential data $\bold{X}$, is squeezed to a direction-wise descriptor $\bold d$ by global average pooling (GAP).
This is prevalent in feature engineering to model multiple output features, which can be replaced by more complex aggregation operations \cite{RF37}.
To be specific, the $n$-th element of $\bold d\in \mathbb{R}^{n},$ is defined by:
\begin{equation}
\label{eq:d}
d_{n} =\frac{1}{K^{2}}  \sum_{i=1}^{K} \sum_{j=1}^{K} V_{n}(i,j),
\end{equation}
furthermore, to utilize the local descriptors of group features, we use a simple gating mechanism with sigmoid activation and two parameterized linear layers for dimensionality reduction and enhancement. 
This is an effective and commonly used operation in feature engineering to model inter-channel correlations \cite{RF37}
\begin{equation}
\label{eq:Translate}
\bold T = sigmoid\left (\bold F_{2} \delta \left (\bold F_{1} \bold d \right )   \right ),
\end{equation}
where, the $\sigma$ is ReLU function, $\bold F_{1}\in \mathbb{R}^{\frac{N}{r} \times N}$, $\bold F_{2}\in \mathbb{R}^{N \times \frac{N}{r}}$, the specific selection of $r$ according to \cite{RF37}

To ensure that multiple directional responses are allowed to be emphasized instead of being one-hot activated.
In this paper, we add the input of DAA back to the output through skip connections.
This provides a shortcut and further improves performance \cite{RF39}.
The output of the DAA module is obtained by recalibrating the $\bold V_{k}$:
\begin{equation}
\label{eq:V}
 \widetilde{\bold  V} =\bold  T\left (\bold V,\left \{ \bold F_{1} ,\bold F_{2}  \right \}  \right ) + \bold V.
\end{equation}
\textbf{Analyze}: The DAA module acts as weights adapted to the filter bank, which is different from the idea used for underground object detection \cite{RF19}, so that the design is not just about preserving the index with the maximum response.
The DAA module intrinsically emphasizes the main directional features and encourages other kernels to generate other types of filters to remove noise/clutter.

\begin{table*}[]
    \centering
        \renewcommand{\arraystretch}{1.1}
\begin{tabular}{c|lll|cc}
\hline
\multicolumn{1}{l|}{\multirow{2}{*}{\textbf{Method/Chanllenge}}} & \multicolumn{3}{c|}{\textbf{GROUNDED}}                                                               & \multicolumn{2}{c}{\textbf{CMU-GPR}}      \\ \cline{2-6} 
\multicolumn{1}{l|}{}                                            & \multicolumn{1}{c}{Sunny/Sunny} & \multicolumn{1}{c}{Sunny/Rainy} & \multicolumn{1}{c|}{Sunny/Snowy} & Nsh\_b              & Gate\_g             \\ \hline
\textbf{3D ResNet18}                                             & 0.94/0.95/0.96                  & 0.49/0.61/0.67                  & 0.46/0.55/0.60                   & 0.46/0.61/0.69 & 0.70/0.89/0.93       \\ \hline
\textbf{3D ResNet50}                                             & 0.93/0.95/0.96                  & 0.50/0.64/0.71                  & 0.47/0.56/0.62                   & 0.53/0.64/0.72 & 0.72/0.91/0.95  \\ \hline
\textbf{3D ResNet101}                                            & 0.93/0.94/0.95                  & 0.48/0.62/0.70                  & 0.45/0.56/0.62                   & 0.48/0.64/0.70 & 0.69/0.87/0.95  \\ \hline
\textbf{CMUNet}                                                  & \multicolumn{1}{c}{0.94/0.95/0.96}           & \multicolumn{1}{c}{0.50/0.61/0.69}           & \multicolumn{1}{c|}{0.46/0.55/0.61}           & 0.43/0.69/0.76 & 0.75/0.93/0.97 \\ \hline
\textbf{DEC}                                                     & \multicolumn{1}{c}{0.40/0.47/0.51}           & \multicolumn{1}{c}{0.21/0.31/0.41}           & \multicolumn{1}{c|}{0.27/0.37/0.46}           & 0.22/0.48/0.54 & 0.23/0.33/0.40 \\ \hline
\textbf{MWSNet}                                                  & \multicolumn{1}{c}{0.92/0.93/0.94}           & \multicolumn{1}{c}{0.48/0.55/0.63}           & \multicolumn{1}{c|}{0.43/0.54/0.67}           & 0.57/0.69/0.76 & 0.81/0.94/0.95 \\ \hline
\textbf{SeqVLAD}                                                 & 0.67/0.86/0.90                  & 0.32/0.43/0.50                 & 0.28/0.40/0.47                   & 0.23/0.27/0.29 & 0.36/0.55/0.65 \\ \hline
\textbf{SeqOT}                                                   & \textbf{0.93/0.95/0.96}         & 0.49/0.58/0.64                  & 0.40/0.44/0.50                   & 0.36/0.54/0.65 & 0.51/0.81/0.86 \\ \hline
\textbf{LGPRNet}                                                 & 0.89/0.92/0.94                  & 0.54/0.60/0.67                  & 0.47/0.57/0.66                   & 0.56/0.65/0.70 & 0.75/0.87/0.90 \\ \hline
\textbf{EDENet}                                                  & \textbf{0.93/0.95/0.96}         & \textbf{0.71/0.81/0.85}         & \textbf{0.57/0.66/0.70}          & \textbf{0.81/0.86/0.89} & \textbf{0.95/0.99/0.99} \\ \hline
\end{tabular}
\caption{Comparison to SOTA methods on benchmark datasets. The best is highlighted in bold (recall@1/5/10).}
\label{diffresult}
\end{table*}

\subsubsection{Shift Invariant Unit}
The above analysis indicates that changes in the underground dielectric constant can cause slight changes in the characteristic positions of significant targets underground.
This module designs a simple CNN cascaded max pooling scheme to further boost the specificity of global features $\widetilde{\bold  V}$ to $\eta = \bold W_{2} * pool\left (\bold W_{1}*\widetilde{\bold  V} +b_{1} \right )  + b_{2}$, while maintaining a certain degree of feature shift invariance, demonstrated in previous studies \cite{RF40}.

\subsection{Feature Aggregation Layer}
In the feature aggregation layer, we convert the output of parallel multi-scale EDEBlocks into a high-level compact descriptor $\bold \eta$ using an MLP. 
The underlying idea of feature aggregation is to shift the focus from local features and cluster center extraction to leveraging the fully connected layers' capacity to holistically combine features \cite{RF38}. 
In this setup, each neuron within the feature aggregation layer can perceive image inputs at different scales, thereby possessing a comprehensive receptive field.
\begin{equation}
\label{eq:eta}
\bold \eta = concat\left ( \eta_{1},\eta_{2},...,\eta_{i} \right ), 
\end{equation}

\begin{equation}
\label{eq:o}
\bold O = \sigma\left (\bold W * \bold \eta + \bold b \right ),
\end{equation}
where $\bold W \in \mathbb{R}^{n\times d}$, $n$ is the dimension of high-level desciptor $\eta$, and $d$ is a manually selected dimension. 
The final output $\bold O$ of the network is flattened and processed by L2-normalized, which is consistent with the other PR \cite{RF38}.

\subsection{Loss Function}
During the training phase, we follow the common practice of PR works \cite{RF28,RF29} to find the nearest reference image for each query in the embedded space as the final positive sample Samples with actual geographic locations exceeding 3 meters are considered negative.

We represent the global embedding features of queries and positive and negative samples as $F_{q}$, $F_{p}$ and $F_{n}$, and the network is trained using max margin triplet loss:
\begin{equation}
\label{eq:loss}
\begin{split}
\mathcal L( F_{q} &,\left \{ F_{p} \right \},\left \{ F_{n}  \right \} ) = \\ &N_{p} \left ( \delta + max\left ( d\left ( F_{q},F_{p} \right )   \right )  \right ) - \sum_{N_{n}}\left ( d \left ( F_{q},F_{n} \right )\right ).  
\end{split}
\end{equation}

\begin{figure*}[htbp]
    \centering
    \includegraphics[width=1\linewidth]{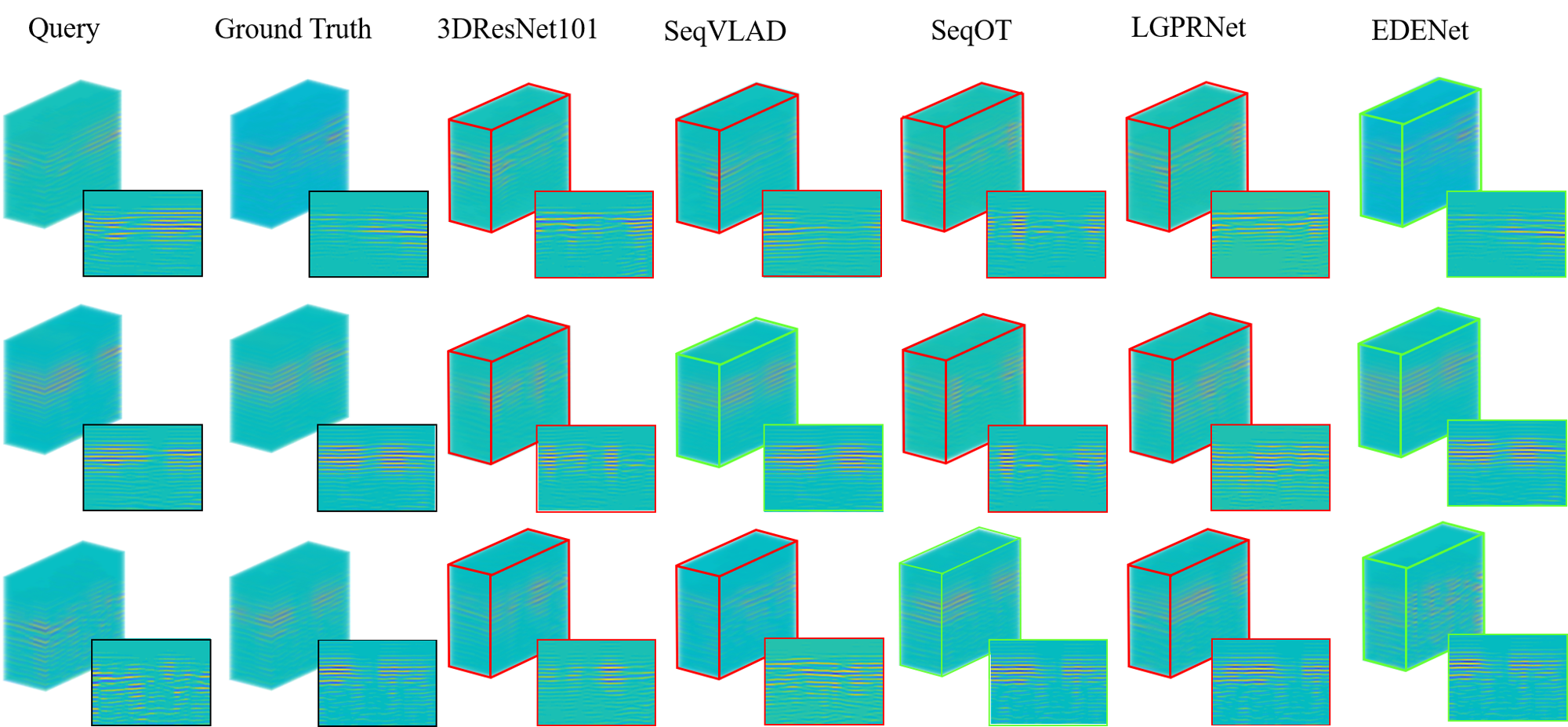}
    \caption{Qualitative results of GROUNDED (Sunny/Rainy). Green and red boxes in indicate correct and wrong match.
    }
\label{MIT}
\end{figure*}

\section{Experiment}

In this section, we demonstrate our approach through several sets of experiments as follows:

\begin{itemize}
    \item  Comparison with state-of-the-art techniques for GPR based methods and representative techniques for sequence localization using camera and LiDAR, which is our main purpose.
    \item  Ablation experiments to demonstrate the effectiveness of each module and combination of the network.
    \item  Evaluation of the model's computational efficiency and number of parameters, which is important for whether the model can be used in actual deployment.
\end{itemize}
All the above experiments were validated on the 3D multi-channel public dataset GROUNDED \cite{RF10} and the 2D single channel CMU-GPR Datasets \cite{RF12}, demonstrating the performance of our proposed method.

\subsection{Implement Detail}

All experiments were conducted on a system equipped with an Intel i7-11700K CPU and an Nvidia RTX 3090 GPU.
\subsubsection{Network Training and Testing}
For the GROUNDED, $run\_0008, 0046, 0047, 0080$ are used for training, while $run\_0054, 0056, 0057, 0090$ served as validation and test sets. 
In the CMU-GPR, Nsh\_b and Gate\_g were used for cross-validation to prevent data leakage. 
The training and validation sets were split with a 7:3 ratio, optimize with Adam\cite{RF45} and learning rate to 0.0001.

\subsubsection{Parameter Setting}
In the GROUNDED dataset, model parameters and descriptor dimensions were aligned with the settings of SeqVLAD \cite{RF21}.
In the CMU-GPR dataset, all descriptor generation dimensions were set to 400 to maintain consistency with the handcrafted DEC features.

\subsection{Performance of EDENet w.r.t. State-of-the-art}

\begin{table}[]
    \centering
    \renewcommand{\arraystretch}{1.2}
\begin{tabular}{c|c|c}
\hline
\multicolumn{1}{l|}{} & GROUNDED                     & CMU-GPR                      \\ \hline
Replacing Gabor with conv & 0.63/0.68        & 0.71/0.78         \\ \hline
DAA module (w/o)      & 0.62/0.70          & 0.74/0.80         \\ \hline
Learnable Gabor + DAA & \textbf{0.71/0.76} & \textbf{0.75/0.82} \\ \hline
\end{tabular}
\caption{Ablation results of learnable Gabor filter and DAA }
\label{Ablation}
\end{table}

The first experiment compares the proposed method against existing techniques. 
Given the scarcity of solutions for GPR-based localization, ResNet series methods are used as baselines, with state-of-the-art (SOTA) methods for cameras and LiDAR also included for comparison. 
These alternative sensor methods are retrained on the GPR dataset as outlined in original papers \cite{RF21, RF22}.

In the GROUNDED dataset, SeqVLAD and SeqOT are selected as SOTA solutions for camera/LiDAR, with feature descriptor parameters set according to \cite{RF21, RF22}. 
For the CMU-GPR dataset, CMUNet, MWSNet and DEC (all designed only for single-channel GPR data) are added as comparatives.

Table \ref{diffresult} demonstrates that the proposed method outperforms others across different scenarios in both datasets. 
The DEC method, which describes energy changes in underground scenes, lacks discriminative power. Camera and LiDAR-based methods like SeqVLAD and SeqOT, which depend on rich detail features, tend to overfit and exhibit poor generalization on GPR sequence data due to high feature texture repetition. 
Additionally, clustering in the feature aggregation stage may overemphasize irrelevant information, limiting performance. 
CMUNet, MWSNet, and LGPRNet, based on ResNet, U-Net, and InceptionNetwork respectively, excel at capturing multi-scale detail textures but are hindered by an excessive focus on new information introduced by changes in underground dielectric constants, as initially analyzed, leading to model bottlenecks.

\begin{figure*}[htbp]
    \centering
    \includegraphics[width=1\linewidth]{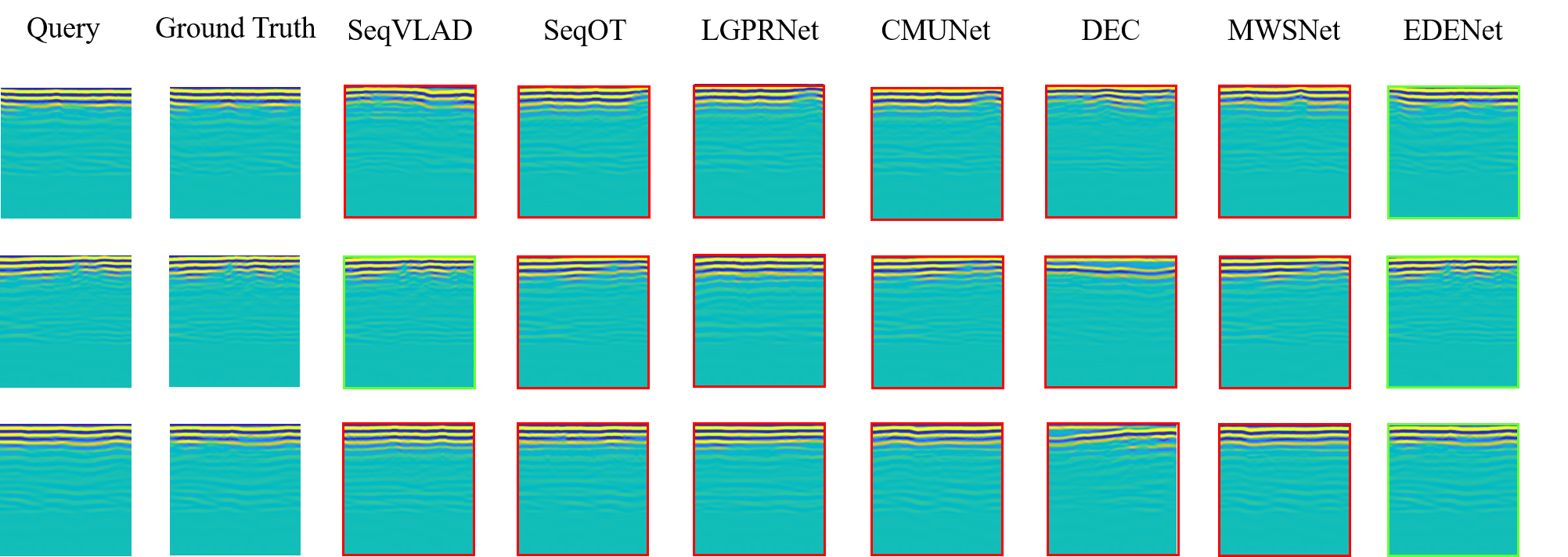}
    \caption{Qualitative results of CMU-GPR.  Green and red boxes in indicate correct and wrong match.
    }
\label{CMU}
\end{figure*}

\subsection{Ablation Experiment}
\subsubsection{EDEBlock}
The primary innovation presented in this paper is the EDEBlock. 
We conducted ablation experiments on the LGF and DAA modules within EDEBlock using the GROUNDED and CMU-GPR datasets, measuring performance in terms of recall@1 and recall@5. 
As shown in Table \ref{combinations}, substituting the learnable Gabor filter with a standard convolutional kernel resulted in an average performance drop of 6$\%$, highlighting the importance of the LGF in capturing directional features. 
Similarly, the DAA module, which is designed as a plug-and-play component, demonstrated its critical role in selectively enhancing the output of the learnable Gabor filter. 
Its removal led to a 4$\%$ drop in performance, underscoring the necessity of adaptive attention in maintaining the integrity of the learned features. 
The combination of LGF and DAA achieved the best results, validating the proposed method for robust underground feature representation through the main directional description.

The qualitative results and dataset comparisons in Fig. \ref{MIT} and Fig. \ref{CMU} further demonstrate the robustness of our method against existing approaches. 
This is particularly evident in the GROUNDED dataset, where varying weather conditions cause fluctuations in the underground dielectric constant, leading to changes and shifts in characteristic amplitude. 
Despite these challenges, our proposed scheme effectively captures echo texture variations and accurately identifies corresponding locations

\begin{table*}[htbp]
    \centering
\begin{tabular}{cccc|lllll}
\hline
\multicolumn{1}{l}{EDE-35} & \multicolumn{1}{l}{EDE-23} & \multicolumn{1}{l}{EDE-11} & \multicolumn{1}{l|}{EDE-5} & $k=8$ & $k=16$ & $k=32$ & $k=64$ & $k=96$ \\ \hline
\checkmark                                   & \checkmark                                   & \checkmark                                   & \checkmark                                   & 69.3         & 72.4          & 71.5          & \textbf{76.0}          & 73.2          \\ \hline
\checkmark                                   &                                     & \checkmark                                   & \checkmark                                   & 65.6         & 72.1          & 72.2          & 74.6 & 72.3          \\ \hline
\checkmark                                   & \checkmark                                   & \checkmark                                   &                                     & 58.7         & 62.0          & 72.6          & 73.2          & 70.4          \\ \hline
                                    & \checkmark                                   & \checkmark                                   & \checkmark                                   & 69.1         & 71.0          & 72.1          & 68.6          & 72.0          \\ \hline
\checkmark                                   &                                     &                                     & \checkmark                                   & 67.2         & 67.0          & 70.1          & 72.6          & 72.0          \\ \hline
\checkmark                                   &                                     & \checkmark                                   &                                     & 66.3         & 65.1          & 68.7          & 70.1          & 71.2          \\ \hline
\end{tabular}
\caption{Recall @5 Recognition Rate (\%) with Different Scale Combinations and hyperparameters $k$ }
\label{combinations}
\end{table*}

\subsubsection{Combination and Parameter of EDENet}

The scale of EDEBlocks is crucial for capturing the diverse spatial features in GPR data. By employing kernels of different sizes—$35\times35$, $23\times23$, $11\times11$, and $5\times5$—EDENet effectively captures both broad and fine details. Larger kernels, like $35\times35$, are proficient at capturing low-frequency, global information essential for understanding the overall underground structure. In contrast, smaller kernels, such as $5\times5$, focus on high-frequency, local details, which are vital for detecting fine-grained features.

The combination of these scales allows EDENet to comprehensively encode GPR data by leveraging the strengths of both large and small kernels. However, introducing additional scales can lead to performance saturation, where the benefit of capturing more information diminishes due to increased redundancy. The learnable Gabor filters, which form the core of EDEBlocks, are not orthogonal and can overlap in the feature space, leading to redundant information that may degrade performance through overfitting.

The parameter $k$ directly impacts the angular resolution of the Gabor filters within each EDEBlock. A larger $k$ increases angular resolution, enabling the Gabor filters to model a broader range of orientations, which is advantageous for GPR-based place recognition as it enhances the model’s ability to differentiate between locations. However, a higher $k$ also adds complexity to the filter bank, increasing the risk of redundancy by producing filters that capture similar features without adding new information.

This analysis underscores the importance of carefully balancing the configuration of EDEBlocks in EDENet. 
The multi-scale kernel combination is essential for capturing the full spectrum of GPR features, but must be chosen to avoid performance saturation. 
Likewise, the parameter $k$ should be optimized to enhance angular resolution while minimizing redundancy.
The selected configuration of $35\times35$, $11\times11$, and $5\times5$ EDEBlocks with $k=64$ was empirically validated as the optimal setup, offering a balanced and effective approach to GPR-based place recognition.

\subsubsection{Performance and Runtime Analysis}

As shown in Table \ref{runtime}, the fourth experiment reports the average runtime and model size of all methods across ten trials.
Our method is the most compact in model size while maintaining competitive performance in parameter count and runtime. 
Notably, it does not require specialized functions for sequence descriptor similarity evaluation, enabling acceleration with the Faiss library \cite{RF23}. 
This results in a runtime of 188 Hz with a 100-frame GPR sequence, surpassing the 126 Hz GPR data acquisition rate and demonstrating the potential for real-time deployment. 
In contrast, other methods with the same descriptor size require more processing time due to the excessive parameters in backbones designed for Camera/LiDAR data, reducing efficiency.

\begin{table}[htbp]
\renewcommand{\arraystretch}{1.1}
\begin{tabular}{c|c|c|c}
\hline
                             & Parameters                         & Model Size     & Runtimes        \\ \hline
3DResNet18            & 1.6×10\textasciicircum{}10          & 846Mb          & 0.006s          \\ \hline
3DResNet50            & 1.71*10\textasciicircum{}10         & 1.24G          & 0.008s          \\ \hline
3DResNet101           & 3.54*10\textasciicircum{}10         & 1.63G          & 0.014s          \\ \hline
SeqVLAD               & 2.06*10\textasciicircum{}9          & \textbf{786Mb}    & \textbf{0.004s} \\ \hline
SeqOT                 & \underline{6.21*10\textasciicircum{}8} & 831Mb          & 0.016s          \\ \hline
LGPRNet               & 3.01*10\textasciicircum{}9          & 1.5G           & 0.011s          \\ \hline
EDENet                & \textbf{5.0*10\textasciicircum{}8}     & \underline{828Mb} & \underline{0.005s}    \\ \hline
\end{tabular}
\caption{Runtime and Model Size}
\label{runtime}
\end{table}

\section{Conclusion}

This paper proposes a novel neural network, EDENet, designed to generate compact descriptors based on GPR sequence echoes for PR tasks.
EDENet introduces a learnable Gabor filter layer and a direction-aware attention mechanism to effectively integrate echo geometric feature encoding while removing redundant information.
The current work primarily focuses on the exploration of the backbone network, with a significant concentration of parameters in the final feature aggregation layer.
Future work will continue to investigate the design of the feature aggregation layer for positioning based on GPR sequence echoes, aiming to achieve a more efficient and lightweight solution.

\section{Acknowledgments}
This work was supported by the National Natural Science Foundation of China ( No. 62303475 and No. 62401576) and the Hunan Province Innovation and Entrepreneurship Project (No. XJZH2024009 and No. XJZH2024010)

\bibliography{aaai25}

\end{document}